\documentclass{article}
\usepackage{PRIMEarxiv}
\usepackage{float}
\usepackage[utf8]{inputenc} 
\usepackage[T1]{fontenc}    
\usepackage{hyperref}       
\usepackage{url}            
\usepackage{booktabs}       
\usepackage{amsfonts}       
\usepackage{nicefrac}       
\usepackage{microtype}      
\usepackage{lipsum}
\usepackage{fancyhdr}       
\usepackage{graphicx}       
\graphicspath{{media/}}     
\usepackage{soul,color}
\usepackage{multirow}
\usepackage{amsmath}
\usepackage{amsbsy}
\usepackage{amssymb}
\usepackage{caption}
\usepackage{subcaption}
\usepackage{subfloat}

\usepackage{bm}
\title{Online Anchor-Based Training for Image Classification Tasks
\thanks{\textit{\underline{Citation}}: 
\textbf{M. Tzelepi, V. Mezaris, "Online Anchor-Based Training for Image Classification Tasks", Proc. IEEE Int. Conf. on Image Processing (ICIP 2024), Abu Dhabi, United Arab Emirates, Oct. 2024, accepted for publication.}}
\thanks{(c) IEEE. This is the authors' accepted version. The final IEEE-published version can be found in IEEE Xplore. }
}

\author{
  Maria Tzelepi and Vasileios Mezaris\\ 
  Information Technologies Institute (ITI) \\
  Centre of Research and Technology Hellas (CERTH) \\
  Thessaloniki, Greece\\
  \texttt{\{mtzelepi,bmezaris\}@iti.gr} \\
}

\begin{document}
\maketitle
\begin{abstract}
In this paper, we aim to improve the performance of a deep learning model towards image classification tasks, proposing a novel anchor-based training methodology, named \textit{Online Anchor-based Training} (OAT). The OAT method, guided by the insights provided in the anchor-based object detection methodologies, instead of learning directly the class labels, proposes to train a model to learn percentage changes of the class labels with respect to defined anchors. We define as anchors the batch centers at the output of the model. Then, during the test phase, the predictions are converted back to the original class label space, and the performance is evaluated. The effectiveness of the OAT method is validated on four datasets.
\end{abstract}


\section{Introduction}\label{sec:intro}

Image classification refers to the task of assigning a class label to an image based on its visual content, and it is a task of vivid research interest since the past few decades \cite{lu2007survey}. In the relevant literature, several training methodologies are employed for addressing image classification tasks. In the conventional supervised approach of the deep learning era \cite{krizhevsky2012imagenet}, a deep neural network accepts the input images and through multiple levels of transformation, an output is produced. The softmax operation is applied to the output producing the probability of each image to belong to each of the classes of the task. The cross entropy loss is commonly used for training a neural network. Deep neural networks of various architectures have been used, including ones of convolutional nature \cite{krizhevsky2012imagenet,rawat2017deep,sun2019evolving}, recurrent nature \cite{dhruv2020image,chandra2017improving} or the more recent attention-based transformers \cite{dosovitskiy2020image,touvron2021going,liu2023survey}.

The cross entropy loss used in the conventional supervised training approach suppresses the similarities of the samples with the classes, encoded at the output distribution, except for the correct class. Nevertheless, these similarities usually reveal useful information beyond the class label, while at the same time training with one-hot class labels can lead to over-fitting. To this end, another training approach, closely related to knowledge distillation methods \cite{zhang2020distilling,tzelepi2021efficient}, utilizes the so-called soft labels. In this training approach, the model is trained both with the class labels (i.e., hard labels) and the soft labels, providing improved generalization ability, compared to the conventional supervised training. 

Another approach, related to the previous one, for improving the training process in order to achieve better performance in terms of accuracy, is the label smoothing \cite{muller2019does,zhang2021delving}. In this approach, the labels are computed as a weighted average of the class labels and the uniform distribution over labels. Finally, another established training approach for improving the performance of a deep learning model towards a vision recognition task is to introduce regularization objectives, including a plethora of methods \cite{kukavcka2017regularization}.

In this work, we propose a novel training methodology for improving the classification performance of a deep learning model, considering image classification tasks. The proposed methodology is inspired by the anchor-based object detection methodologies \cite{ren2015faster,liu2016ssd,he2017mask,lin2017focal,li2019dynamic}. More specifically, the target of object detection is to predict the bounding boxes of objects of interest. To achieve this goal, the anchor-based methods propose to train the network in order to learn offsets instead of learning directly the absolute coordinates. That is, the network is provided with predefined boxes, known as anchors, and the objective is re-formulated as prediction of offsets. The underpinning idea behind this approach is that is easier for the network to learn offsets. For example, the SSD object detector \cite{liu2016ssd}, instead of learning the ground truth coordinates of the bounding box, learns the logarithm of the ratio between the ground truth coordinates and the anchor's coordinates.

The anchor-based training idea has also been utilized for addressing other tasks rather than object detection. For example, in a recent work, an anchor-based approach has been proposed for facial landmark localization, proposing to regress the offsets against each of the predefined anchors \cite{xu2021anchorface}. Furthermore, an anchor-based approach has been recently proposed for addressing time-series forecasting tasks \cite{tzelepi2023improving}. In this work, the authors deal with the electric load demand forecasting task, proposing, instead of directly predicting the electricity demand of the target day, to train a model for predicting percentage changes of the load of the target day with respect to an anchor, using as anchor the electric load one week prior the target day.

In this paper, we aim to introduce this idea, which is well-suited to regression tasks, to more challenging image classification tasks. That is, we propose a novel anchor-based training approach for image classification tasks, named \textit{Online Anchor-based Training} (OAT). More specifically, similarly to the aforementioned methodologies, we propose to dynamically define the anchors, and instead of learning the class labels, to learn percentage changes of them with respect to the aforementioned anchors. That is, the problem is transformed into predicting the offsets instead of predicting the class labels. We propose to compute the anchors, in an online fashion, as the batch centers at the output layer of the model.

The proposed online anchor-based training methodology can be applied to deep learning models regardless of their complexity (e.g., from simple lightweight models to complex transformers), remarkably enhancing their performance, as it is experimentally validated. Furthermore, as it also experimentally validated, the proposed OAT methodology has similar computational cost to the conventional supervised process, both in terms of training and inference times. It should be, finally, highlighted that, in this paper, our goal is improve the classification performance of deep learning models towards the image classification tasks, proposing a novel training methodology, rather than a specified model architecture. That is, the OAT method is orthogonal to existing models and it could be readily combined with most existing ones for further improving their performance.

The remainder of the manuscript is organized as follows. Section \ref{sec:method} presents in detail the proposed online anchor-based training methodology for image classification tasks. Subsequently, Section \ref{sec:exp} provides the experimental evaluation of the proposed method, followed by the conclusions in Section \ref{sec:con}.

\section{Proposed Method}\label{sec:method}

\begin{figure*}[!h]
  \centering
    \includegraphics[width=0.8\textwidth]{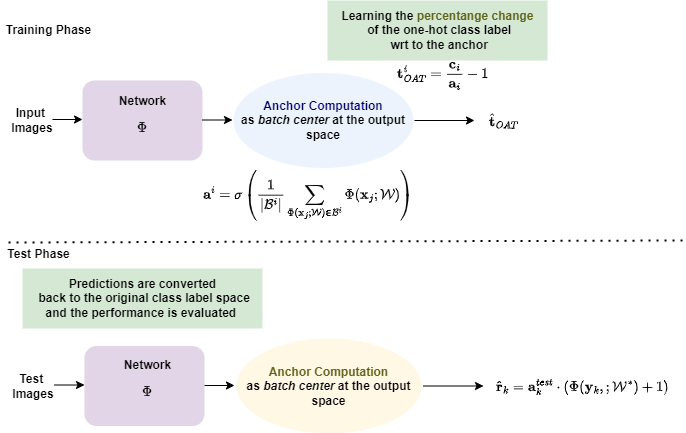}
        \caption{Training and test phases of the proposed OAT method: In the training phase we dynamically compute the anchors at the output space of the model, and train the network to learn percentage changes of the class labels w.r.t to the anchors. Then during the test phase, the predictions are converted back to the original class label space, so that the accuracy of the model in predicting the correct class labels can be evaluated.}
        \label{fig:oat}
\end{figure*}

In this paper, we propose a method, named \textit{Online Anchor-based Training} (OAT), for image classification tasks, motivated by the insights provided in the anchor-based object detection methods \cite{ren2015faster,liu2016ssd}. Specifically, we propose to define anchors, and instead of learning the class labels, to learn percentage changes of the class labels with respect to the anchors. As anchors, we propose to use the batch centers at the output layer of the network, in an online fashion, followed by the softmax operation in order to transform them into values between 0 and 1, summing to 1. 

More specifically, we consider a $C$-class classification task, and the labeled data ${\{\mathbf{x}_i,\mathbf{c}_i \}}_{i=1}^{N}$, where $\mathbf{x}_i\in\Re^{D}$ is an input vector and $D$ its dimensionality, and $\mathbf{c}_i \in  \mathcal{Z}^C$ corresponds to its $C$-dimensional one-hot class label. For an input space $\mathcal{X} \subseteq \Re^{D}$ and an output space $ \mathcal{F} \subseteq \Re^{C}$, we consider as $\Phi(\cdot\, ; \mathcal{W}):\mathcal{X} \rightarrow \mathcal{F}$ a deep neural network, with set of parameters $\mathcal{W}$, which transforms its input vector to a $C$-dimensional vector. That is, $\Phi(\mathbf{x}_i\, ; \mathcal{W}) \in \mathcal{F}$ corresponds to the output vector of $\mathbf{x}_i \in \mathcal{X}$ given by the network $\Phi$ with parameters $\mathcal{W}$.

The anchor, $\mathbf{a}^i \in\Re^{C}$, for a sample $i$, inside the batch $\mathcal{B}^i$ is computed online throughout the training process as follows:  
\begin{equation}
\mathbf{a}^i =  \sigma\left( \frac{1}{|\mathcal{B}^i|} \sum_{\Phi(\mathbf{x}_j ; \mathcal{W}) \in \mathcal{B}^i} \Phi(\mathbf{x}_j;\mathcal{W})\right),
\end{equation}
where $\sigma(\cdot)$ is the softmax operation.

Thus, the targets for the proposed OAT training method are formulated as follows for a sample $i$: 
\begin{equation}\label{eq:transform}
    \mathbf{t}_{OAT}^i = \frac{\mathbf{c}_i}{\mathbf{a}^i} - 1. 
\end{equation}

We train the network with the aforementioned targets using Mean Squared Error (MSE):
\begin{equation}\label{eq:mse}
    \mathcal{J} = \frac{1}{N}\sum_1^N (\mathbf{t}_{OAT}^i - \hat{\mathbf{t}}_{OAT}^i)^2.
\end{equation}

Thus, by optimizing eq. (\ref{eq:mse}), the network $\Phi( \cdot \, ; \mathcal{W}^*)$ is trained to predict a percentage change according to eq. (\ref{eq:transform}). Then, during the test phase, considering the corresponding set of test samples, ${\{\mathbf{y}_k,\mathbf{r}_k \}}_{k=1}^{K}$, where $\mathbf{y}_k\in\Re^{D}$, and $\mathbf{r}_k \in  \mathcal{Z}^C$, the inverse transformation is performed, so that the predictions for the test samples are converted back to the original space of the class labels, in order to evaluate the accuracy of the method in predicting the correct labels. Based on the fundamental assumption of supervised learning that training and test samples are drawn from the same distribution, we use as anchor in test phase the corresponding batch center at the output of the model denoted as $\mathbf{a}^{test}_k$, in order to perform the aforementioned transformation. This is formulated as follows: 
\begin{equation}
    \hat{\mathbf{r}}_k = \mathbf{a}^{test}_k \cdot  (\Phi(\mathbf{y}_k, ; \mathcal{W}^*)+1),
\end{equation}
where $\Phi(\mathbf{y}_k, ; \mathcal{W}^*)=\hat{\mathbf{t}}_{OAT}^k$.
In this way, the class label predictions of the proposed OAT method are obtained, and the performance of method can be evaluated and be compared with the conventional approaches. The training and test phases of the proposed OAT method are illustrated in Fig. \ref{fig:oat}. It should be finally noted that a limitation of the proposed method is that it uses multiple test samples to compute the anchors in the test phase. However, it should be highlighted that this does not limit the method to be applied only to batches of test samples, and instead, it can be applied even for a single test sample, as it is also experimentally validated. In this case, the batch center is identical to the single sample.

\section{Experimental Evaluation}\label{sec:exp}
In this paper, we have conducted experiments on four datasets for evaluating the performance of the proposed anchor-based training method. The datasets' and the networks' descriptions, along with the evaluation metrics and implementation details, and finally the experimental results are provided in the subsequent subsections.

\subsection{Datasets}
We use four datasets to evaluate the effectiveness of the proposed OAT method, i.e., Cifar-10 \cite{krizhevsky2009learning}, UCF-101 \cite{soomro2012ucf101}, Event Recognition in Aerial videos (ERA) \cite{eradataset}, and Biased Action Recognition (BAR) \cite{nam2020learning}. The used datasets vary in terms of size, ranging from 1,473 to 50,000 training images, in terms of number of classes, ranging from 6 to 101 classes, and also in terms of image resolution, including images of size $32\times32$ and $640\times640$.

More specifically, Cifar-10 consists of 50,000 train images and 10,000 test images divided into 10 classes. Images are of size $32\times32$. UCF-101 is an action recognition data set of 13,320 YouTube action videos, consisting of 101 action categories. The middle frame is derived from each video, and the train and test sets are formed according to the original splits \cite{soomro2012ucf101}. Thus, the train set contains 9,537 images, while the test set contains 3,783 images. Images are of size $320\times240$. ERA dataset is an event recognition in unconstrained aerial videos. It consists of 1,473 train images and 1,391 test images, divided into 25 event classes. Images are of size $640\times640$. BAR is a real-world image dataset with six action classes which are biased to distinct places. The dataset consists of 1,941 train images and 654 test images. All images of the BAR dataset are over $400\times300$. Images in all the latter three datasets are resized to $224\times224$.

\subsection{Network Architectures}
Network architectures of varying complexity are utilized in all the used datasets. More specifically, in the Cifar-10 dataset, two networks for accepting input images of size $32\times32$ are utilized, in order to avoid image resizing and interpolation by $\times7$. That is, a simple lightweight model, consisting of two convolutional layers with six and sixteen kernels of size $5\times5$ respectively and three fully connected layers (128 × 64 × 10), and a heavyweight modified Wide-ResNet-28-10 model (abbreviated as WRN-28-10). In the rest three datasets, we use ResNet-18 \cite{he2016deep} Wide-ResNet-50-2 \cite{zagoruyko2016wide} (abbreviated as WRN-50-2), and VIT-L-16 \cite{dosovitskiy2020image}, pretrained on ImageNet weights. A linear layer is added to the output of the networks, with neurons equivalent to the number of the classes of each dataset.

\subsection{Evaluation Metrics}
Test accuracy is used to evaluate the effectiveness of the proposed training methodology. Training and test time in terms of seconds are also reported.

\subsection{Implementation Details}
The proposed OAT method was implemented using the PyTorch framework. The networks are trained using the mini-batch gradient descent, with a mini-batch of 32 samples. The learning rate (lr) is set to 0.001, and the momentum is 0.9. The networks are trained for 100 epochs on an NVIDIA GeForce RTX 3090 with 24 GB of GPU memory. 

It should be emphasized that the above lr is optimal for both the compared approaches, i.e., training with and without OAT. To this aim, we also provide representative evaluation results on the UCF-101 dataset, using the ResNet-18 model for different values of lr.  The results are presented in Table \ref{tab:lr}. Best performance is printed in bold considering the two compared methods of training with and without OAT, while best performance of each method is underlined. As it is shown, the used lr is the optimal for both the methods. Note also that for lr=0.01 the model using the proposed methodology, the model does not converge to a good solution.

Furthermore, since the anchor computation is linked with the batches, we also perform experiments on the UCF-101 dataset, using the ResNet-18 model, for various batch sizes. The experimental results are provided in Table \ref{tab:batch}. As it is demonstrated the proposed OAT method provides better performance compared to training without OAT for each considered case. 

Finally, it should be noted that the proposed method does not technically require to pre-train the model with the conventional supervised approach on the dataset of interest, using the original class labels, however, due to the fact that the anchors are dynamically computed, we experimentally noticed that such a pre-training for only a few epochs provides significant performance boost. To this aim, in all the conducted experiments, we have conventionally pre-trained the models for only 10 epochs, and thus to ensure the fairness of the comparisons, we train the model with the OAT method for 90 epochs. Furthermore, in Table \ref{tab:anchor}, we provide representative experimental results on the UCF-101 dataset, for various epochs of conventional pre-training, where it is evident that the proposed OAT method provides steadily enhanced performance. 

\begin{table}[!ht]
\begin{center}
\caption{UCF-101: Test accuracy for various lrs using the ResNet-18 model \cite{he2016deep}.} \label{tab:lr}
\begin{tabular}{|c|c|c|c|}
  \hline
  & \multicolumn{3}{|c|}{\bf{Learning Rate}}\\ \hline
  \bf{Method} & 0.01 & 0.001 & 0.0001 \\ \hline
  W/o OAT & \bf{70.288} & \underline{72.403} & 70.579 \\ \hline
  OAT & 59.397 & \underline{\bf{74.253}} & \bf{71.980} \\
  \hline
\end{tabular}
\end{center}
\end{table}

\begin{table}[!ht]
\begin{center}
\caption{UCF-101: Test accuracy for various batch sizes using the ResNet-18 model \cite{he2016deep}.} \label{tab:batch}
\begin{tabular}{|c|c|c|c|}
 \hline
   & \multicolumn{3}{|c|}{\bf{Batch Size}}\\ \hline
  \bf{Method} & 32 & 64 & 128 \\ \hline
  W/o OAT & 72.403 & 71.557 & 72.112 \\ \hline
  OAT & \underline{\bf{74.253}} & \bf{72.932} & \bf{72.852} \\
  \hline
\end{tabular}
\end{center}
\end{table}

\begin{table}[!ht] 
\begin{center}
\caption{UCF-101: Test accuracy for different epochs of pre-training using the ResNet-18 model \cite{he2016deep} (W/o OAT: 72.403).} \label{tab:anchor}
\begin{tabular}{|c|c|}
  \hline
  \bf{Epoch} & \bf{Test Accuracy} 
  \\
  \hline
  W/o pre-training & \bf{72.641} \\ \hline 
  5 & \bf{73.117} \\ \hline
  10 & \underline{\bf{74.253}} \\ \hline
  20 & \bf{73.222} \\
  \hline
\end{tabular}
\end{center}
\end{table}

\subsection{Experimental Results}
In this section, we provide the experimental results of the conducted experiments on the four utilized datasets. We compare the performance of the proposed OAT method, in terms of test accuracy, against the training of a model of identical architecture without OAT, i.e., with the original class labels using the conventional cross entropy loss. Best performance is printed in bold.

First, in Table \ref{tab:cifar}, the experimental results on Cifar-10 dataset are provided, for both the lightweight and the relatively heavyweight models. As it can be shown, the proposed OAT methodology achieves considerably improved performance against training without OAT in all the considered cases, validating our claim that the proposed method can improve the classification performance for models of different complexity. 

Subsequently, in Table \ref{tab:ucf} the corresponding experimental results on the UCF-101 dataset are provided, for the different used model architectures. As it can be demonstrated, the OAT method achieves superior performance over the compared approach in all the considered cases. Correspondingly, in Table \ref{tab:era} the evaluation results on the ERA dataset are presented, providing consistently improved performance over the conventional supervised training. It should also be noted that the OAT method applied to the VIT-L-16 model achieves superior performance over the current state-of-the-art (\cite{eradataset}). In Table \ref{tab:bar} the corresponding results on the BAR dataset are presented, where the OAT method significantly improves the models' performance. Superior performance over the current state-of-the-art (\cite{li2022discover}) is also obtained with the OAT methodology using both the WRN-50-2 and VIT-L-16 models. 

\begin{figure*}
\begin{minipage}[b]{0.32\linewidth}
\centering
\includegraphics[width=\textwidth]{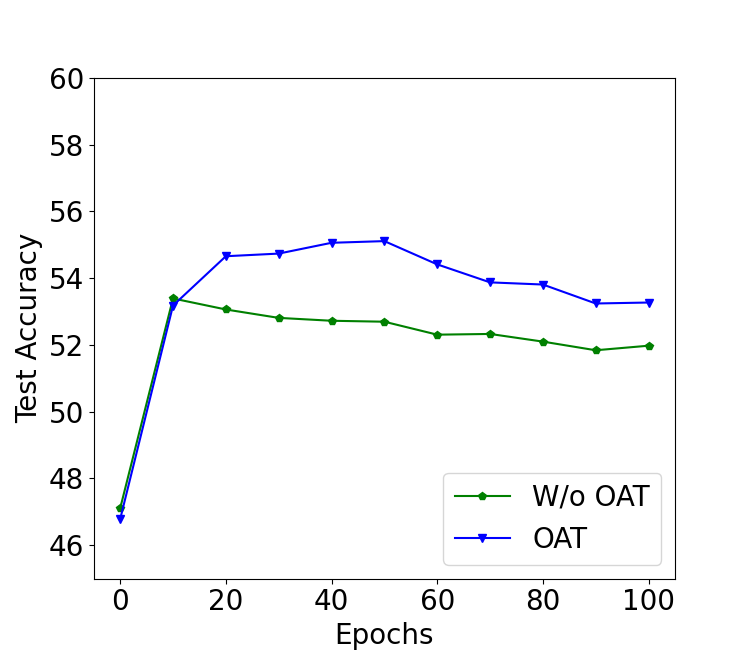}
\caption{ERA dataset: Test accuracy throughout the training epochs using the ResNet-18 model.}
\label{fig:era1}
\end{minipage}
\hspace{0.1cm}
\begin{minipage}[b]{0.32\linewidth}
\centering
\includegraphics[width=\textwidth]{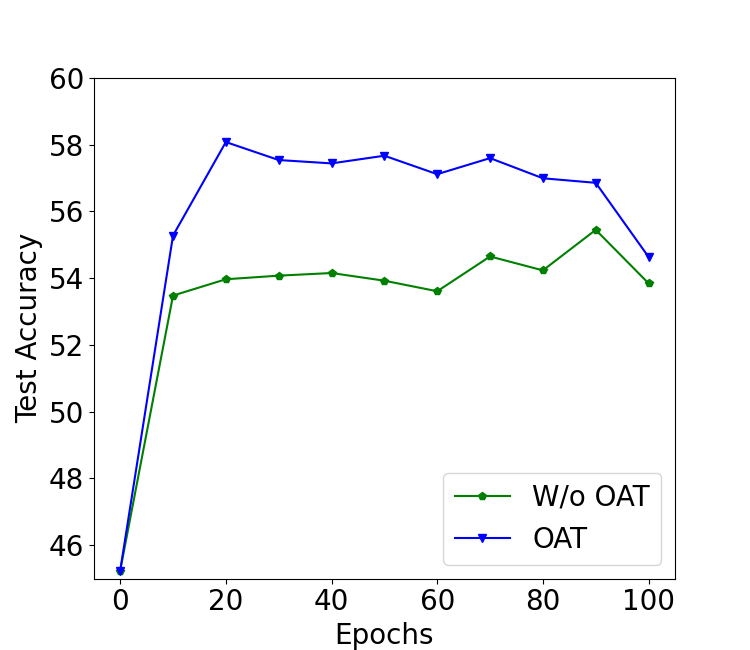}
\caption{ERA dataset: Test accuracy throughout the training epochs using the WRN-50-2 model.}
\label{fig:era2}
\end{minipage}
\hspace{0.1cm}
\begin{minipage}[b]{0.32\linewidth}
\centering
\includegraphics[width=\textwidth]{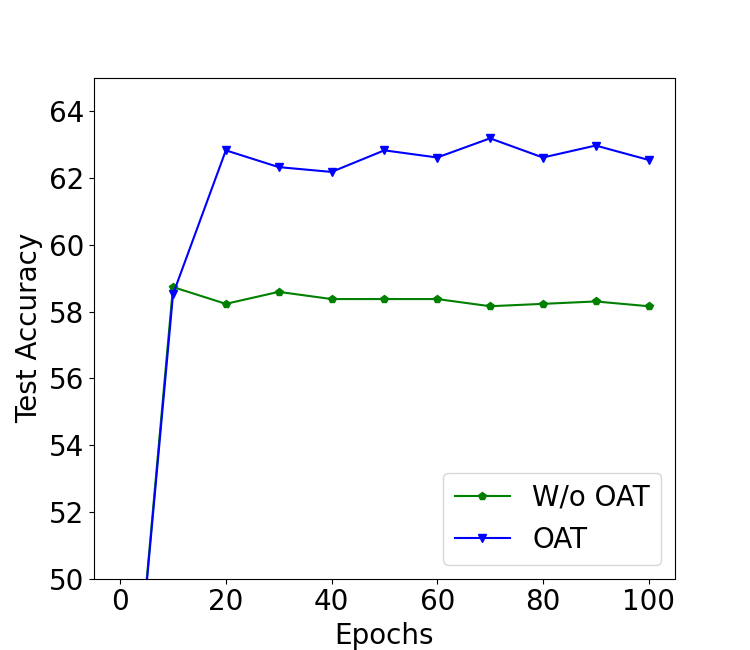}
\caption{ERA dataset: Test accuracy throughout the training epochs using the VIT-L-16 model.}
\label{fig:era3}
\end{minipage}
\end{figure*}

\begin{figure}[!ht]
\begin{minipage}[b]{0.5\linewidth}
\includegraphics[width=\textwidth]{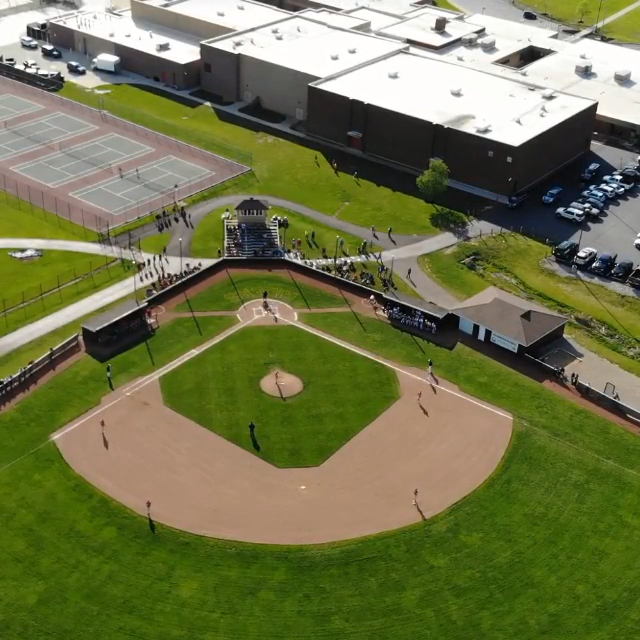}
\subcaption{Class: \textit{Baseball} }
\end{minipage}
\begin{minipage}[b]{0.5\linewidth}
\includegraphics[width=\textwidth]{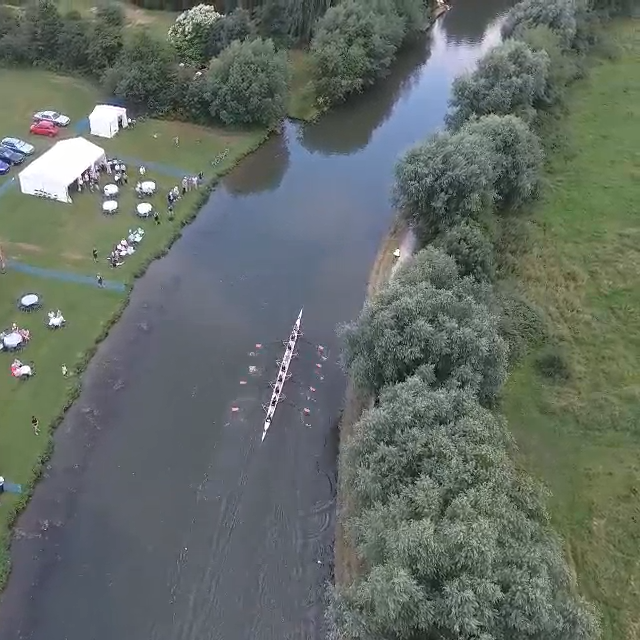}
\subcaption{Class: \textit{Boating} }
\end{minipage}
\caption{ERA dataset: Examples of misclassified test images without the OAT method. Both the images where classified to the class \textit{Non-event} using the model trained w/o OAT, while they were correctly classified as \textit{Baseball} and \textit{Boating} respectively, using the proposed OAT method.}
\label{fig:examples}
\end{figure}

Furthermore, in Figs. \ref{fig:era1}, \ref{fig:era2}, and \ref{fig:era3} we indicatively present the test accuracy throughout the training epochs using the ResNet-18, WRN-50-2, and VIT-L-16 models respectively, for the two approaches of training with and without the proposed OAT method on the ERA dataset. The curves have been smoothed for better comparison. As it is illustrated, the proposed method provides steadily better performance over the training epochs. Finally, in Fig. \ref{fig:examples} some examples of misclassified images without the OAT method, while correctly classified using the proposed OAT method, are presented.

\begin{table}[!ht]
\begin{center}
\caption{Cifar-10: Test accuracy for the proposed OAT method against training w/o OAT, using both the utilized models.} \label{tab:cifar}
\begin{tabular}{|c|c|c|}
  \hline
  \bf{Method} & \bf{Lightweight} & \bf{WRN-28-10 \cite{zagoruyko2016wide}}  \\ \hline
  W/o OAT & 65.660 & 89.130  \\ \hline
  OAT & \bf{66.540} & \bf{92.020}  \\
  \hline
\end{tabular}
\end{center}
\end{table}

\begin{table}[!ht]
\begin{center}
\caption{Test accuracy for the proposed OAT method against training w/o OAT, using three different models on the UCF-101 dataset.} \label{tab:ucf}
\resizebox{0.455\textwidth}{!}{%
\begin{tabular}{|c|c|c|c|}
  \hline
  \bf{Method} & \bf{ResNet-18 \cite{he2016deep}} & \bf{WRN-50-2 \cite{zagoruyko2016wide}} & \bf{VIT-L-16 \cite{dosovitskiy2020image}}  \\ \hline
  W/o OAT & 72.403 & 78.853  & 78.324\\ \hline
  OAT & \bf{74.253} & \bf{79.038}  & \bf{79.170}\\ \hline
\end{tabular}}
\end{center}
\end{table}

\begin{table}[!ht]
\begin{center}
\caption{Test accuracy for the proposed OAT method against training w/o OAT, using three different models on the ERA dataset.} \label{tab:era}
\resizebox{0.455\textwidth}{!}{%
\begin{tabular}{|c|c|c|c|}
  \hline
  \bf{Method} & \bf{ResNet-18 \cite{he2016deep}} & \bf{WRN-50-2 \cite{zagoruyko2016wide}} & \bf{VIT-L-16 \cite{dosovitskiy2020image}}  \\ \hline
  W/o OAT & 54.565 & 56.506  & 58.879\\ \hline
  OAT & \bf{57.565} & \bf{59.597}  & \bf{63.767}\\ \hline
\end{tabular}}
\end{center}
\end{table}

\begin{table}[!ht]
\begin{center}
\caption{Test accuracy for the proposed OAT method against training w/o OAT, using three different models on the BAR dataset.} \label{tab:bar}
  \resizebox{0.455\textwidth}{!}{%
\begin{tabular}{|c|c|c|c|}
  \hline
  \bf{Method} & \bf{ResNet-18 \cite{he2016deep}} & \bf{WRN-50-2 \cite{zagoruyko2016wide}} & \bf{VIT-L-16 \cite{dosovitskiy2020image}}  \\ \hline
  W/o OAT & 61.621 & 70.183  & 75.688\\ \hline
  OAT & \bf{64.526} & \bf{73.853}  & \bf{77.676}\\ \hline
\end{tabular}}
\end{center}
\end{table}

Furthermore, as it was previously mentioned the above process can be applied even for a single test sample. In this case the batch center is identical to the single sample. To validate the effectiveness of the OAT method under such a scenario, we provide the corresponding experimental results on the UCF-101 dataset, where the OAT achieves test accuracy 73.407, against training w/o OAT which achieves test accuracy 72.403.   

Finally, regarding the computational complexity of the proposed anchor-based training methodology, the OAT method's complexity is similar to the conventional supervised training, both in terms training and inference time. More specifically, training the model to learn changes is slightly faster than the conventional supervised approach. In addition, the online anchor computation does not affect the process, since it consists in a simple computation at the output space, which is in general low-dimensional. For example, considering the UCF-101 dataset, which is the more computationally demanding out of the utilized datasets, due to the number of classes (and, in turn, the dimension of the output representation), the training time per epoch is 13.956 secs, while 1 epoch of conventional training takes 14.943 secs. Considering the test time, the whole evaluation takes 4.391 secs considering the OAT methodology, while it takes 4.219 secs considering the conventional supervised training approach.

It should also be highlighted that the online nature of the proposed anchor-based method is considerably faster compared to an approach that would pre-compute the anchors. That is, in a preliminary investigation in which we computed the anchors for each sample based on its class label, in the test phase we had to find the nearest anchor to each test sample in order not to exploit class label information introducing an additional computational cost. More specifically, considering the UCF-101 dataset, the whole evaluation considering the aforementioned anchor-based approach takes 9.116 secs, while as mentioned, the OAT evaluation takes 4.391 secs.  

\section{Conclusions}\label{sec:con}
In this paper, we deal with image classification tasks, aiming to improve the performance of a deep learning model, proposing a novel online anchor-based training pipeline. The OAT methodology proposes to train a model to learn percentage changes of the one-hot class labels with respect to defined anchors, rather than learning directly the class labels. The batch centers at the output of the model are used as anchors. Then during the test phase, the predictions are converted back to the original class label space, and the performance is evaluated. The experimental evaluation on four datasets, using models of varying complexity, validates the effectiveness of the OAT methodology.

\section*{Acknowledgements}
This work has been funded by the European Union as part of the Horizon Europe Framework Program, under grant agreement 101070109 (TransMIXR).

\end{document}